\pdfoutput=1

\documentclass[11pt]{article}

\usepackage{acl}

\usepackage[T1]{fontenc}

\usepackage[utf8]{inputenc}
\usepackage{microtype}
\usepackage{times}
\usepackage{latexsym}
\usepackage{amsmath,amssymb,amsfonts}
\usepackage{algorithmic}
\usepackage{graphicx}
\usepackage{textcomp}
\usepackage{xcolor}
\usepackage{hyperref,enumitem,xcolor}
\usepackage{caption}
\usepackage{subcaption}
\usepackage{booktabs}
\newcommand{\Lab}{\mathcal{L}}
\title{Fine-Tuning Pretrained Language Models With Label Attention for Biomedical Text Classification}

\author{Bruce Nguyen and Shaoxiong Ji\\
  Aalto University, Finland \\
  \texttt{\{bruce.nguyen,~shaoxiong.ji\}@aalto.fi}}

\begin{document}
\maketitle
\begin{abstract}
The massive scale and growth of textual biomedical data have made its indexing and classification increasingly important. 
However, existing research on this topic mainly utilized convolutional and recurrent neural networks, which generally achieve inferior performance than the novel transformers.
On the other hand, systems that apply transformers only focus on the target documents, overlooking the rich semantic information that label descriptions contain.
To address this gap, we develop a transformer-based biomedical text classifier that considers label information.
The system achieves this with a label attention module incorporated into the fine-tuning process of pretrained language models (PTMs).
Our results on two public medical datasets show that the proposed fine-tuning scheme outperforms the vanilla PTMs and state-of-the-art models.

\end{abstract}

\section{Introduction}

Clinical practices and biomedical research generate a vast and rapidly growing amount of textual data \citep{dash2019big}. 
This includes written information about patients in electronic health records (EHRs) and millions of articles recorded by biomedical research databases.
To analyze this magnitude of data effectively, we need large-scale organization and classification \citep{botsis_nguyen_woo_markatou_ball_2011}.
However, such schemes (e.g., medical coding, medical subject headings (MeSH) indexing) mainly employ specially educated professionals for manual classification.
Not only is this approach expensive, but it has also been shown to be unreliable \citep{dai2020fullmesh, omalley_cook_price_wildes_hurdle_ashton_2005}.
As a result, automated medical text indexing and classification systems have been an active research area in recent years. 

\begin{figure}[t]
    \centering
    \includegraphics[width=\columnwidth]{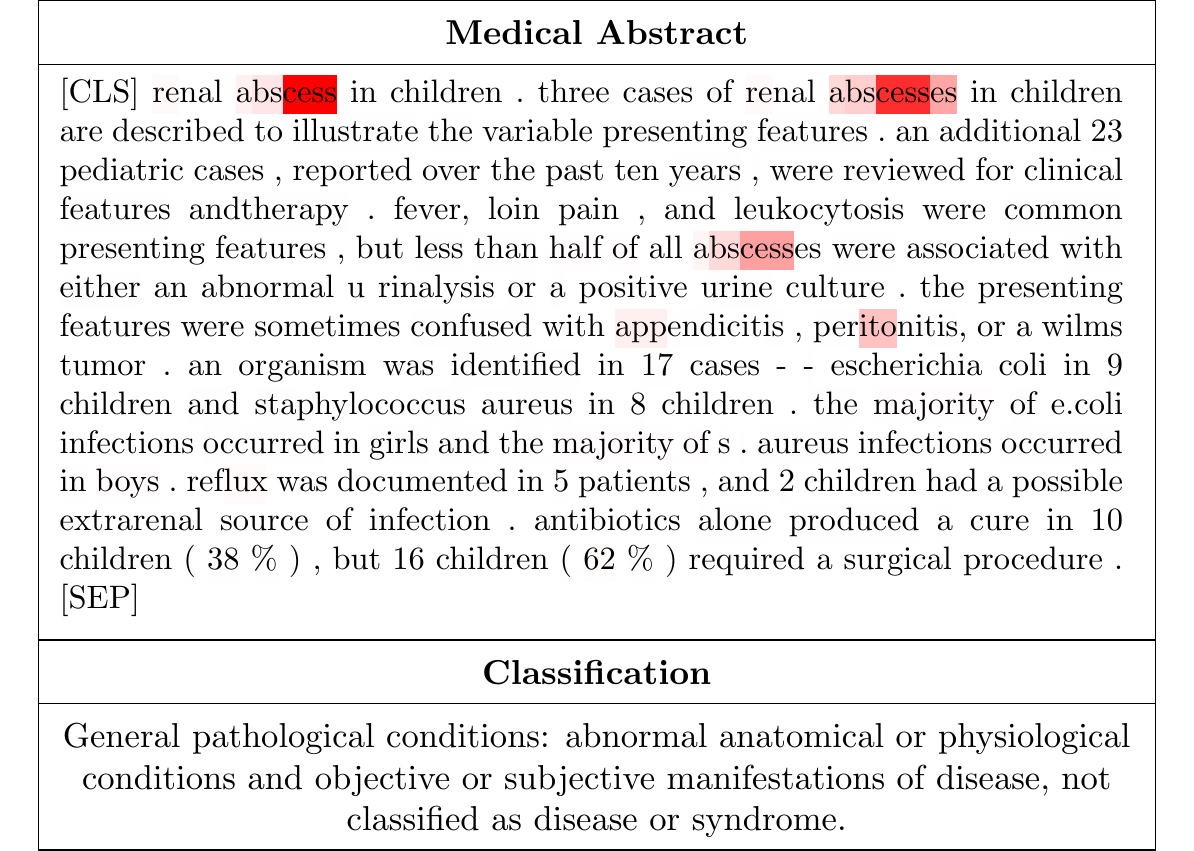}
    \caption{
    A sample of the Disease-5 dataset classified by LAME with attention scores visualized. We can see that the model attends heavily to the word `abscess', which is appropriate evidence that the text is about `general pathological conditions'.
    }
    \label{fig:explanation}
\end{figure}

Recent works in this area have primarily focused on convolutional and recurrent neural network-based architecture and word embeddings (e.g., word2vec, fastText, and ELMo).
For example, the ML-NET \citep{du2019ml} comprises an RNN-based architecture with ELMo embeddings and a label count prediction module. 
CAML \citep{mullenbach2018explainable} utilizes the word2vec embedding and CNNs for document encoding and the dot-product label attention mechanism for injecting label information.
Compared to such architectures, transformer-based pretrained language models (PTMs) such as BERT and GPT-3 have achieved great performance gains, dominating leaderboards in most natural language processing tasks.
However, studies that apply PTMs based on transformers in bioinformatics have mainly worked on improving the pretraining process with better data and techniques \citep{wu_roberts_datta_du_ji_si_soni_wang_wei_xiang_etat_2019}. 
This forgoes the valuable information labels contain, especially in the medical context.

In this paper, we propose a novel neural network architecture that involves fine-tuning biomedical versions of the BERT language model \citep{Devlin2019BERTPO,biobert} with label attention for explainable biomedical text classification. 
We name our 
method \textbf{L}abel \textbf{A}ttention \textbf{M}odeling for \textbf{E}xplainability (LAME)\footnote{The source code will be made available.}.
Our motivations for the design are three-fold: 
(1) BERT, a state-of-the-art (SOTA) transformer-based PTM, generates a more accurate representation of biomedical texts; 
(2) fine-tuning it, instead of performing feature extraction, leads to better results, as shown by \citet{peters2019tune}; 
(3) label attention helps attend to n-grams in documents relevant to the label. 
Moreover, the label attention module can explain the prediction via attention scores, thus giving us insights into the system's decision-making process.
This interpretability is critical in medical sciences, differentiating our model from most deep learning approaches.
We conduct experiments on two public biomedical datasets and show that our approach outperforms the conventionally fine-tuned BERT, a feature extraction-based BERT with label attention, and the SOTAs in both instances.

\begin{figure*}
    \centering
    \includegraphics[width=\textwidth]{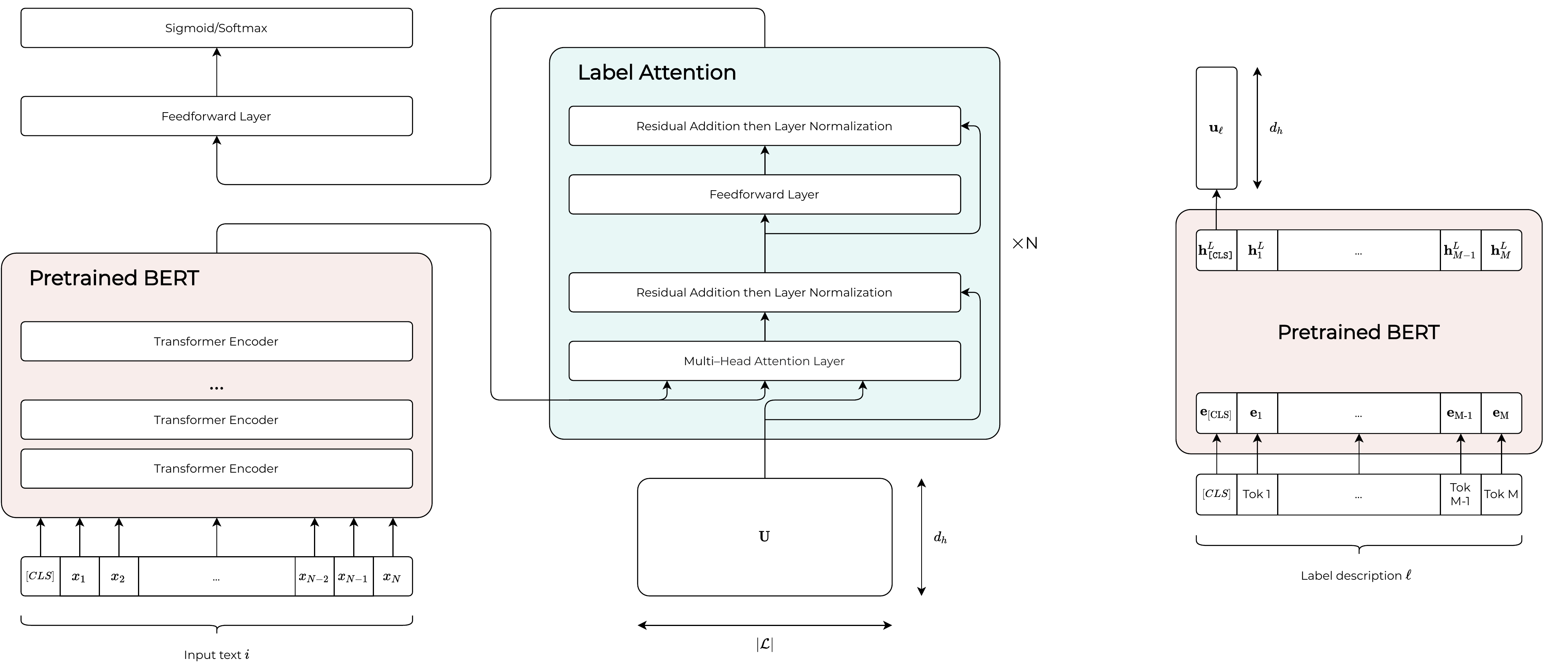}
    \caption{
    Left: The illustration of our proposed LAME architecture with pretrained contextualized models and label attention mechanism.
    Right: The label embedding process with the final hidden representation $\mathbf{h}^{L}_{\texttt{[CLS]}}$ of the special \texttt{[CLS]} token.
    }
    \label{fig:mlem}
\end{figure*}

\section{Method}
This paper formulates the LAME fine-tuning architecture for the multi-class and multi-label classification settings.
Let the set of all labels be  $\Lab$. Given the input sequence $\mathbf{x}$, the model determines the probability of class or label $y_\ell$ for all $\ell \in \Lab$ based on the $\mathbf{u}_{\ell}$ label description representation.
Figure \ref{fig:mlem} shows the design of both the model and the label description embedding process.
In detail, the target sequence and the label descriptions are first passed into the pretrained BERT model, resulting in deep contextualized representations.
Then, our label attention layer helps the label descriptions attend to important locations within the sequence.
The output is aggregated and normalized with a feedforward layer and a softmax/sigmoid function to get the final class/label probabilities.

\subsection{Fine-tuning biomedical BERT}

BERT, introduced by \citet{Devlin2019BERTPO}, generates deeply bidirectional contextualized representations of language.
It achieves this with a stacked Transformer encoder structure \cite{vaswani2017attention} and advanced pretraining tasks such as masked language modeling and next sentence prediction.
To adapt to different domains, BERT can be flexibly fine-tuned or pretrained on in-domain textual data.
We choose to fine-tune the BioBERT \cite{biobert} variant---which was specifically created for biomedical NLP---because of its empirical performance on the experimented datasets. 
We refer to it in our design only as BERT for the convenience of notation.

As the input of BERT in the model, we have the target document sequence $\mathbf{x} = \{x_1, x_2, \text{...}, x_N\}$, where $N$ is the length of the document.
Let $\mathbf{H}^{l}$ be the representation created at the $l^{\text{th}}$ Transformer encoder layer of BERT, where $l \in [1, L]$, and $d_h = 768$ is the hidden dimension hyperparameter.
The process can be formulated as follow:
\begin{equation*}
    \small\mathbf{H}^{l} = \text{TransformerEncoder}_l(\textbf{H}^{l-1})\text{,}
\end{equation*}
where $\mathbf{H}^{0}$ is the input embeddings of $\mathbf{x}$.
The output is the matrix representation $\mathbf{H}^{L} \in \mathbb{R}^{N \times d_h}$ of the target document sequence.
We call it by the short-hand notation $\mathbf{H}$.

\subsection{Label Description Embedding}
For each label $\ell$, we input its description into the pretrained BERT to obtain its representation vector $u_{\ell} \in \mathbb{R}^{d_h}$. This is done by taking the final hidden state $\mathbf{h}^{L}_{\texttt{[CLS]}}$ of the special \texttt{[CLS]} token for classification tasks.
We repeat this step for every training instance $i$ and thus fine-tune the model's parameters concerning the label representation.
As a result, the pretrained BERT jointly learns the most appropriate representations for both the document $\mathbf{x}$ and the label $\ell$.
All the label representation vector $\mathbf{u}_{\ell}$ with $\ell \in \Lab$ is then concatenated into matrix $\mathbf{U} \in \mathbb{R}^{\vert \Lab \vert \times d_h}$  for further processing by the label attention layer.

\subsection{Label Attention Layer}
Our design of this layer is inspired by the Transformer model architecture \cite{vaswani2017attention} and its immense success in modeling language dependency.
The layer comprises a multi-head attention sub-layer and a feedforward network, with residual connections \cite{he2016deep} and layer normalization schemes \cite{ba2016layer} after each sub-layer.
We choose multi-head attention instead of conventional attention modules because we hypothesize that it is more effective in high-dimensional space of $\mathbf{H} \in \mathbb{R}^{N \times d_h}$ and $\mathbf{U} \in \mathbb{R}^{\vert \Lab \vert \times d_h}$.
In addition, it is possible to stack blocks of label attention to produce a deeper cross attention signal between the target document and the labels.

In details, we have the output of each sub-layer as $\text{LayerNorm}(\mathbf{x} + \text{Sublayer}(\mathbf{x}))$ where $\mathbf{x}$ is the input and $\text{Sublayer}$ is the operation performed by each sub-layer (multi-head attention and fully connected network).
At the base of the module, the target document representation matrix $\mathbf{H}$ and the label embedding space $\mathbf{U}$ are fed into the multi-head attention layer.
The former represent the key and value and the latter the query for the subsequent scaled dot-product attention calculation:
\begin{align*}
\small
    \mathbf{O}
    = \text{Attention}(\mathbf{U},\mathbf{H},\mathbf{H}) 
    = \text{softmax}(\frac{\mathbf{U}\mathbf{H}^{T}}{\sqrt{d_h}})\mathbf{H} \text{,}
\end{align*}
where $\mathbf{O} \in \mathbb{R}^{\vert \Lab \vert \times d_h}$ is the attention output.
Since we are using multi-head attention, the actual underlying operations are as follows:
\begin{align*}
\small
    & \text{head}_i = \text{Attention}(\mathbf{U}\mathbf{W}^{{Q}}_i, \mathbf{H}\mathbf{W}^{{K}}_i, \mathbf{H}\mathbf{W}^{{V}}_i) \text{,} \\
    & \text{MultiHead}(\mathbf{U},\mathbf{H},\mathbf{H}) = \text{Concat}(
    \text{head}_i)\mathbf{W}^{{O}}
\end{align*}
for all $i \in [1,h]$, where $\mathbf{W}_i$ and $\mathbf{W}^{{O}}$ are the projection parameter matrices.
Regarding the number of heads, we adopt $h=12$ based on analyses by \cite{voita2019analyzing} and our experimental results.
After calculating attention scores, the attention output $\mathbf{O}$ is aggregated with $\mathbf{U}$ and normalized.
The resulting matrix is then passed onto the next sub-layer, keeping its dimensions unchanged throughout the process.

\subsection{Classification \& Loss Function}
Let $\mathbf{O} \in \mathbb{R}^{\vert \Lab \vert \times d_h}$ be the final output of the label attention layer(s).
It is then further fed into another feedforward network and finally, the softmax (for multi-class classification) or sigmoid (for multi-label classification) function to obtain the final label probabilities $y_{\ell}$ for all $\ell \in \Lab$.

{
\renewcommand{\arraystretch}{1.25}
\begin{table}[h]
\small
\centering
\caption{The results of various models on the HoC dataset. LA represents the label attention module.}
\begin{tabular}{lccc} 
\toprule
Model                       & F1             & Precision & Recall  \\
\midrule
CNN~\citep{Kim2014ConvolutionalNN}                          & 80.90           & 84.30 & 77.80            \\
ML-NET \citep{du2019ml}                      & 82.90 & 81.10          & 84.80  \\  

Fine-tuned BERT              & 83.04          & 80.71     & 85.84   \\
Frozen BERT + LA      & 71.61          & 66.33     & 78.74   \\
LAME~ & \textbf{83.30} & 78.84     & \textbf{88.63}   \\
\bottomrule
\end{tabular}
\end{table}
}

{
\renewcommand{\arraystretch}{1.25}
\begin{table}[h]
\centering
\small
\caption{The results of various models on the Disease-5 dataset}
\begin{tabular}{lc} 
\toprule
Model                   & Accuracy         \\ 
\midrule
CNN~\citep{Kim2014ConvolutionalNN}                     & 62.50            \\
BiForest \cite{Wang2019MultiGranularTE} & 65.20  \\ 

Fine-tuned BERT               & 82.91            \\
Frozen BERT + LA & 83.18            \\
LAME                    & \textbf{83.65}   \\
\bottomrule
\end{tabular}
\end{table}
}

For biomedical multi-class text classification, we employ the prevalent cross-entropy loss. 
Unfortunately, in the multi-label case, there is often a substantial degree of imbalance within the datasets, making the commonly-used binary cross-entropy loss suboptimal \cite{eban2017scalable}.
Therefore, we also experiment with a loss function based on the F-measure for mini-batch training:
\begin{align*}
\small
L_{\text{F-measure}} = \frac{1}{2}(\frac{2\text{tp}}{2\text{tp} + (\text{fp}+\text{fn})} + \frac{2\text{tn}}{2\text{tn} + (\text{fp}+\text{fn})})
\end{align*}
with tp, tn, fp, fn being the true positive, true negative, false positive, and false negative micro-weighted rates within a mini-batch. 
In our experiments, we find that while this loss does not improve the fine-tuning process of base BERT, it helps with the convergence of LAME.

\section{Experiments}

\subsection{Datasets \& Baselines}
We train LAME following the setup of previous SOTA methods on two biomedical datasets: 
(1) Hallmarks of Cancers (HoC)
\footnote{https://github.com/sb895/Hallmarks-of-Cancer} 
\cite{Baker2016CancerHT}: 
    a multi-label dataset with 1,580 PubMed abstracts labeled with the 10 hallmarks of cancer. 
    We split this dataset into train/dev/test sets with a ratio of 7:1:2.
(2) Medical abstracts of diseases (Disease-5) \footnote{https://github.com/SnehaVM/Medical-Text-Classification--MachineLearning}: a multi-class dataset with 14,102 medical abstracts falling into 5 classes.
    We split the dataset into train/dev/test sets with a ratio of 8:1:1. Since the label descriptions of Disease-5 are not given, we create our own according to our best knowledge.

We take into account previous works while also building our baselines for comparisons.
For HoC and Disease-5, we include the ML-NET and the BiForest model \cite{Wang2019MultiGranularTE} as the SOTA, respectively.
The latter used a tree-structured LSTM and the attention mechanism for more explainable predictions.
The authors in both studies also presented a single-layer one-dimensional CNN baseline \cite{Kim2014ConvolutionalNN}, which we also utilize for comparisons.
In addition to LAME, we experiment with a typically fine-tuned BERT for text classification \cite{sun2019fine}, and a feature extraction-based approach to BERT (frozen BERT) \cite{peters2019tune}, which has a CNN as the classification head, combined with the label attention module. Unfortunately, we find that stacking label attention does not yield performance gains, possibly because of the small size of both datasets. Therefore, their performances are not included in the study.

\subsection{Hyperparameters \& Evaluations}
We experiment with a batch size of 42, 1 and 4 NVIDIA A100 GPUs for the HoC and Disease-5, respectively. 
A slanted triangular learning rate \cite{howard2018universal} with 0.1 warm-up portions is used for BERT and label attention with a 5e-05 and 4e-02 respective maximum learning rate.
The classification head's learning rate, on the other hand, stays unchanged at 1e-03 for the whole 30 epochs of training.
The optimizer used is AdamW \cite{Loshchilov2019DecoupledWD}.
We use micro-weighted F-score, precision, and recall for evaluation in the HoC task, while performance in Disease-5 is only evaluated using accuracy, following \citet{Wang2019MultiGranularTE}.

\subsection{Results}
Results show that our model achieves a noticeable improvement over all the baselines.
Most significantly, LAME outperforms conventionally fine-tuned BERT by approximately 0.3 in F1 in the HoC and 0.5 in accuracy level in the Disease-5 task, showing that injecting label information into the fine-tuning process improves performance.
At the same time, the fine-tuned transformer-based methods are shown to be superior to feature extraction-based ones, with the success of fine-tuned BERT variants.

\subsection{Interpretability}
Given an example $\mathbf{x}$ and an output prediction $\ell$, we can extract the attention scores $\boldsymbol{\alpha}_{\ell}$ generated by the multi-head attention layer within the label attention network.
By aligning this vector $\boldsymbol{\alpha}_{\ell}$ with the input list of tokens, we find the tokens that influence the decision of the model the most.
Figure \ref{fig:explanation} shows an example of the results.
Since BERT employs the WordPiece tokenizer, out-of-vocabulary or rare tokens are divided into subwords. Therefore, the attention can be on such subwords, as evident in the figure.
Readers interested in quantitative interpretability analyses of the label attention module can refer to the extensive study by \citet{mullenbach2018explainable}.

\section{Conclusion}
This paper devises a new deep learning architecture to solve the biomedical text classification task.
This approach involves fine-tuning a pretrained BERT language model with a label attention module.
Therefore, it can learn rich representations of input texts while accounting for the label description information.
Results show the effectiveness of this approach against many baselines and existing methods.
Furthermore, we can interpret the model's decisions thanks to the attention mechanism.
In doing this, we hope to further the use of deep learning in bioinformatics by shedding light on these powerful black-box neural networks.

\bibliography{custom}

\end{document}